\documentclass{article}
\usepackage[utf8]{inputenc}
\usepackage[left=1in, right=1in, top=1in, bottom=1in]{geometry} 
\usepackage{amsmath,amssymb,amsfonts}
\usepackage{url}
\usepackage{hyperref}
\usepackage{graphicx}
\usepackage{csquotes}
\usepackage{pdfpages}
\usepackage{caption}
\usepackage{subcaption}
\usepackage[printonlyused]{acronym}
\usepackage{todonotes}
\usepackage{tikz}
\usepackage{pgfplots}\pgfplotsset{compat=1.15}
\usetikzlibrary{arrows,positioning,calc}
\usepackage{adjustbox}
\usepackage{array}

\newcolumntype{R}[2]{%
    >{\adjustbox{angle=#1,lap=\width-(#2)}\bgroup}%
    b{1em}%
    <{\egroup}%
}
\newcolumntype{L}[2]{%
    >{\adjustbox{angle=#1,lap=(#2)-\width,raise=1.25cm}\bgroup}%
    b{1em}%
    <{\egroup}%
}
\title{The Digital Twin Landscape at the Crossroads of Predictive Maintenance, Machine Learning and Physics Based Modeling}
\author{Brian Kunzer \and Mario Berges \and Artur Dubrawski \thanks{The authors are with Carnegie Mellon University. Copyright \copyright{2022} Carnegie Mellon University.}}

\begin{document}

\maketitle

\abstract{The concept of a digital twin has exploded in popularity over the past decade, yet confusion around its plurality of definitions, its novelty as a new technology, and its practical applicability still exists, all despite numerous reviews, surveys, and press releases. The history of the term digital twin is explored, as well as its initial context in the fields of product life cycle management, asset maintenance, and equipment fleet management, operations, and planning. A definition for a minimally viable framework to utilize a digital twin is also provided based on seven essential elements. A brief tour through DT applications and industries where DT methods are employed is also outlined. The application of a digital twin framework is highlighted in the field of predictive maintenance, and its extensions utilizing machine learning and physics based modeling. Employing the combination of machine learning and physics based modeling to form hybrid digital twin frameworks, may synergistically alleviate the shortcomings of each method when used in isolation. Key challenges of implementing digital twin models in practice are additionally discussed. As digital twin technology experiences rapid growth and as it matures, its great promise to substantially enhance tools and solutions for intelligent upkeep of complex equipment, are expected to materialize. 
}


\section{Introduction}
\begin{figure}[ht]
    \centering
    \includegraphics[width=0.8\textwidth]{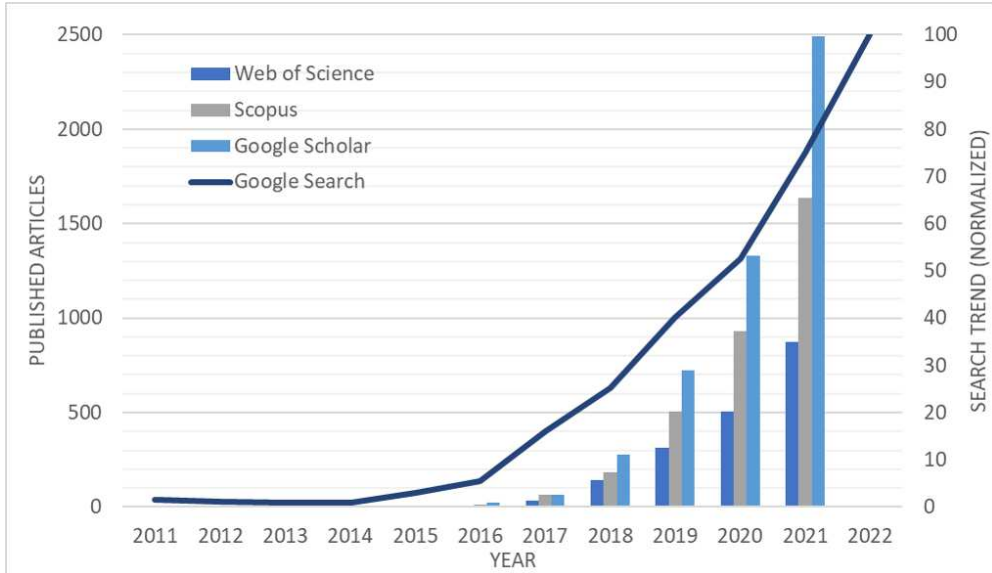}
    \caption{Trends of the query term {\em digital twin} over approximately the last decade show exponential growth. Hit results from three sample databases (Web of Science\texttrademark, Scopus\texttrademark, and Google Scholar\texttrademark) that have published articles with titles containing the terms {\em digital twin} or {\em digital twins}. Line indicates Google\texttrademark~   search (worldwide) frequency for the topic of {\em digital twin}.}
    
    \label{fig:DT_Trend_Analysis}
\end{figure}

The concept of the {\em digital twin} (DT) has been increasingly mentioned over the last decade in both academic and industrial circles. The frequency of a web search topic of {\em digital twin} (includes similar search terms such as, {\em digital twins},  {\em digital twin definition}, {\em what is a digital twin}, etc.) has seen an approximately exponential rise in roughly the past decade (see Figure \ref{fig:DT_Trend_Analysis}). Publication of scholarly articles shows similar trends across several databases including Web of Science\texttrademark, Scopus\texttrademark, and Google Scholar\texttrademark. Yet, the definition of what a digital twin consists of has evolved since its initial introduction, for better or for worse, with some attaching various adjectives to broaden it, as well as others insisting on inclusion of tangential topics to stake novelty claims to the idea in specific technical fields. 

This manuscript seeks to provide clarity on defining {\em digital twin}, through exploring the history of the term, its initial context in the fields of product life cycle management (PLM), asset maintenance, and fleet of equipment management, operations, and planning. A definition for a minimally viable digital twin framework is also provided to alleviate ambiguity of the DT. Furthermore, a brief tour through its applications across industries will be provided to clarify the context in which digital twins are used today. Thereafter, the application of digital twins in the fields of predictive maintenance, machine learning, and physics based modeling with a keen eye towards their intersections will be investigated. Finally, the challenges of implementing digital twins will be examined. The reader should gain a clear understanding of what is a digital twin, how is it applied, and what obstacles to its use lie ahead.

\subsection{Defining Digital Twin and Related Terms}

\subsubsection{Digital Twin Definitions}
At first glance, the term {\em digital twin} conjures the idea of a simple model, a scaled replica or mathematical expression that is a representation of the physically real object or system, and that, certainly, is not a new concept. Over a century ago, Buckingham~\cite{buckingham1914physically} made use of dimensional analysis to lay the groundwork for engineers and physicists to use scaled physical models in place of the larger physical models they represented. These smaller {\em physical twins} of the larger physical object, such as a ship, an aircraft, a bridge or a building, allowed for testing of designs without having to recreate a full scale model, simply by utilizing equal dimensionless parameters invariant to scale, for example, the Reynolds number when analyzing fluid flow past the object. Decades later, with the advent of modern computers, these physical models were supplemented with computational (digital) models that were digitally drawn with computer aided design (CAD) while using governing equations that could be solved by discretizing the model design space into small volumes, elements, and/or nodes~\cite{Marinkovic2019FEM}. So is it relevant to ask if these design models also fall into the definition of a digital twin?

There are multiple definitions for the concept of DT which vary across different industries, and that can broadly encompass sophisticated computational physics-based models of parts, machine learning algorithms applied to recorded sensor data, CAD models, a repository for part and asset manufacturing and maintenance history, and/or scaled virtual reality environments (see Negri et al.~\cite{negri2017review} for a comprehensive list of different DT definitions); however the common theme connecting these definitions is that a virtual representation of a real physical system, machine, or product, spanning its entire life cycle, is created in order to track changes, provide traceability of parts and/or software, and typically connect embedded sensors and Internet of Things (IoT) devices with databases to document life cycle of the subject item. The term digital twin can also be synonymous with {\em digital database}; however,  although potentially existing in a database, a digital twin can refer to the represented model or simulations (or their aggregate thereof) of the physical object or system.  The first comprehensive definition of {\em digital twin} is often credited to NASA's 2012 {\em Modeling, Simulation, Information  Technology \& Processing Roadmap} ~\cite{shafto2012modeling}, defined therein as: 

\begin{quote}
A digital twin is an integrated multiphysics, multiscale simulation of a vehicle or system that uses the best available physical models, sensor updates, fleet history, etc., to mirror the life of its corresponding flying twin. The digital twin is ultra-realistic and may consider one or more important and interdependent vehicle systems, including propulsion/energy storage, avionics, life support, vehicle structure, thermal management/TPS [thermal protection system], etc. Manufacturing anomalies that may affect the vehicle may also be explicitly considered. In addition to the backbone of high-fidelity physical models, the digital twin integrates sensor data from the vehicle’s on-board integrated vehicle health management (IVHM) system, maintenance history and all available historical/fleet data obtained using data mining and text mining. By combining all of this information, the digital twin continuously forecasts the health of the vehicle/ system, the remaining useful life and the probability of mission success. The systems on board the digital twin are also capable of mitigating damage or degradation by recommending changes in mission profile to increase both the life span and the probability of mission success.~\cite{shafto2012modeling}
\end{quote}

Thus, in its initial definition, the digital twin concept was applied to the service life (planning, maintenance, and operation) of a complex asset, such as an aerospace or astronautical vehicle with thousands, if not millions, of individual parts assembled into a dense physical web of interacting functional systems. A similar DT concept was also proposed by Tuegel et al. for an aircraft, specifically~\cite{tuegel2011reengineering}. Such complex assets are subject to multi-year environmental degradation, requiring sustained maintenance, part replacement, mission dependent equipment swaps, and operational planning including fleet management, service downtime scheduling, and part and personnel logistics. Also accompanying the complex asset is a wealth of data, generated from onboard sensors. This data is often generated in the context of feedback for various control logic applications, but also for condition monitoring, warning indicators, and system alarms. These maintenance requirements and the amount of diagnostic data available align directly with the goals of predictive maintenance (PMx, sometimes also abbreviated PdM): the promise that analyzing and interpreting asset data will allow anticipation of the need for corrective maintenance, its convenient scheduling, and preventing equipment failures~\cite{errandonea2020digital,carvalho2019systematic,miller_system-level_2020}. 

\begin{figure}[ht]
    \centering
    \includegraphics[width=0.7\textwidth]{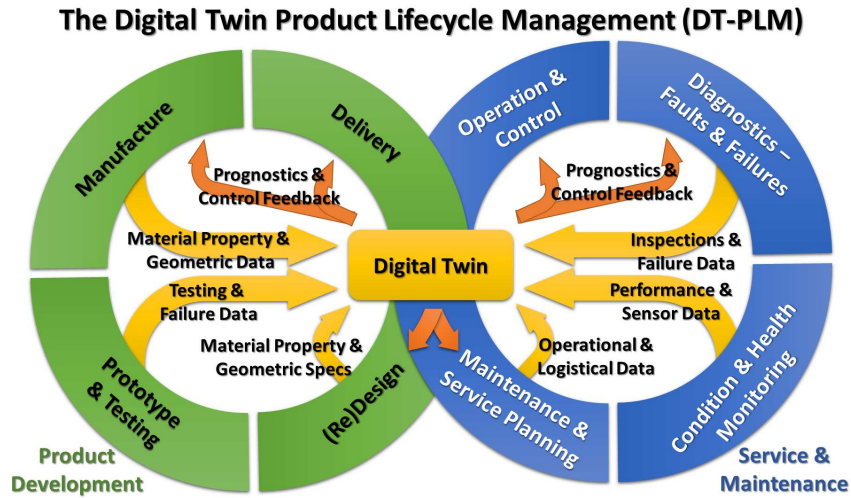}
    \caption{The digital twin product lifecycle management (DT-PLM) concept. The two cycles of product development and product service and maintenance are interwoven. The digital twin model provides a seamless link in connecting these interrelated cycles by providing a common database to store designs, models, simulations, test data, as-manufactured data, algorithms, usage data, and other available information tracked over time and throughout each cycle as the physical twin asset moves along the cycles.}
    
    \label{fig:DT-PLM}
\end{figure}

Another early conceptual framework for digital twins is heavily based on the concept of product lifecycle management (PLM): a systematic approach to managing the series of changes a product goes through, from its design and development, to its ultimate retirement, generational redesign, or disposal~\cite{terzi2010product}. PLM can be visualized as two interwoven cycles of product development, on the one hand, and product service and maintenance, on the other; the former taking place before an asset is used in the real world design intent, and the latter referring to the asset's service support during applicable use (Figure \ref{fig:DT-PLM}). The digital twin can then be conceptualized as a seamless link connecting the interrelated cycles by providing a common database to store designs, models, simulations, algorithms, data, and information tracked over time and throughout each cycle. A claim is made from Grieves~\cite{grieves2016origins} that the following definition preceded the previously mentioned digital twin definition by about a decade under the term {\em Mirrored Spaces Model} which is closely tied with the idea of PLM:

\begin{quote}
The Digital Twin is a set of virtual information constructs that fully describes a potential or actual physical manufactured product from the micro atomic level to the macro geometrical level. The Digital Twin would be used for predicting future behavior and performance of the physical product. At the Prototype stage, the prediction would be of the behavior of the designed product with components that vary between its high and low tolerances in order to ascertain that the as-designed product met the proposed requirements. In the Instance stage, the prediction would be a specific instance of a specific physical product that incorporated actual components and component history. The predictive performance would be based from current point in the product's lifecycle at its current state and move forward. Multiple instances of the product could be aggregated to provide a range of possible future states. Digital Twin Instances could be interrogated for the current and past histories. Irrespective of where their physical counterpart resided in the world, individual instances could be interrogated for their current system state: fuel amount, throttle settings, geographical location, structure stress, or any other characteristic that was instrumented. Multiple instances of products would provide data that would be correlated for predicting future states. For example, correlating component sensor readings with subsequent failures of that component would result in an alert of possible component failure being generated when that sensor pattern was reported. The aggregate of actual failures could provide Bayesian probabilities for predictive uses.  \cite{grieves2016origins}
\end{quote}

In this context, the DT concept is a common realization that most modern machines, systems, and products are generating data from network connected sensors and their design and development typically span multiple engineering domains, as they include mechanical and structural hardware, electronics, embedded software, network communication, and often more. A significant design change or even a part replacement with nonidentical tolerances made in any of the engineering domains will necessitate and propagate significant design changes or changes in performance in other domains. A digital twin provides the connection between prior and future design and operations, to alleviate many communication and logistical problems arising from the intersection of disparate engineering and technical fields, which may exist during a product’s or system’s life cycle. Furthermore, the digital twin’s ability to be updated in real-time or near real-time from embedded and peripheral sensors, allows for deeper analysis of the performance and maintenance of the physical twin asset. 

Rosen et al.~\cite{rosen_about_2015} describe the digital twin concept as the future of a trend in modeling and simulations that has evolved from individual solutions of specific problems to model based systems engineering (MSBE) with multi-physics and multi scale modeling, and then finally to seamless digital twin modeling, linked to operational data throughout the entire life cycle of an asset.  Kritzinger et al.~\cite{kritzinger2018digital} describe levels of automation that distinguish traditional models from digital twin models; digital models have manual data analysis from the physical object and optional manual decision making from the digital object; digital shadows have automated digital threads connecting physical object data to the digital object but still have manual information flow from the digital object to the physical object; only the digital twin has automated data and information flow in each direction via the digital thread to and from each of the physical and digital objects.  A comprehensive review~\cite{Qamsane2020digital} reveals at least ten different definitions for digital twins. A survey of 150 engineers asking what they thought a DT was, revealed the most popular definition was "a virtual replica of the physical asset which can be used to monitor and evaluate its performance" and found that its most useful application was thought to be in the field of predictive maintenance~\cite{Hamer2018feasibility}.

The variety and expanding inclusion of concepts may appear to define DT as a concept equivalent to a traditional {\em simulation} or even {\em data analysis} of sensor data; however, the novelty, in comparison to the 20th century definitions of similar terms, lies in two areas: a) the development of large, connected streams or {\em threads} of sensor data that may be analyzed to improve the understanding of the current system or machine (also a defining feature of {\em Industry 4.0}~\cite{lasi2014industry}); 
and b) the integration of multiple models describing the form, function and behavior of the physical system and its components at various scales and using diverse modeling paradigms.

\subsubsection{Digital Thread, Industry 4.0, Digital Engineering}
Although sometimes used interchangeably with {\em digital twin}, {\em digital thread} has a more specific meaning:

\begin{quote}
The digital thread refers to the communication framework that allows a connected data flow and integrated view of the asset’s data throughout its lifecycle across traditionally siloed functional perspectives. The digital thread concept raises the bar for delivering ``the right information to the right place at the right time'' \cite{leiva2016demystifying}
\end{quote}

Digital thread is often considered synonymous with the {\em connected supply chain} that ``collects data on all parts, kits, processes, machines and tools in real time and then stores that data, enabling digital twins and complete traceability.'' \cite{leiva2016demystifying} Digital thread is also closely related to the concept of {\em Industry 4.0}:

\begin{quote}
The increasing digitalization of all manufacturing and manufacturing-supporting tools is resulting in the registration of an increasing amount of actor- and sensor-data which can support functions of control and analysis. Digital processes evolve as a result of the likewise increased networking of technical components and, in conjunction with the increase of the digitalization of produced goods and services, they lead to completely digitalized environments. Those are in turn driving forces for new technologies such as simulation, digital protection, and virtual or augmented reality. \cite{lasi2014industry}
\end{quote}

Yet another closely related term is {\em digital engineering}, which can be defined as:

\begin{quote}
Digital engineering is defined as an integrated digital approach that uses authoritative sources of systems data and models as a continuum across disciplines to support lifecycle activities from concept through disposal. Digital engineering applies similar emphasis on the continuous evolution of artifacts and the organizational integration of people and process throughout a development organization and integration team [\dots] Digital engineering is sometimes referred to as ``digital thread,'' which is understood as a more encompassing term, though the novelty of both terms has created some dispute over their exact overlap. Both digital thread and digital engineering are an extension of product lifecycle management, a common practice in private industry that involves the creation and storage of a system's lifecycle artifacts, in digital form, and which can be modified as a system evolves throughout its lifecycle. Digital thread and digital engineering both involve a single source of truth, referred to as the authoritative source of truth (ASoT), which contain artifacts maintained in a single repository, and stakeholders work from the same models rather than copies of models. \cite{Shepard2020digitalengineering}
\end{quote}

\begin{figure}[ht]
    \centering
    \includegraphics[width=0.8\textwidth]{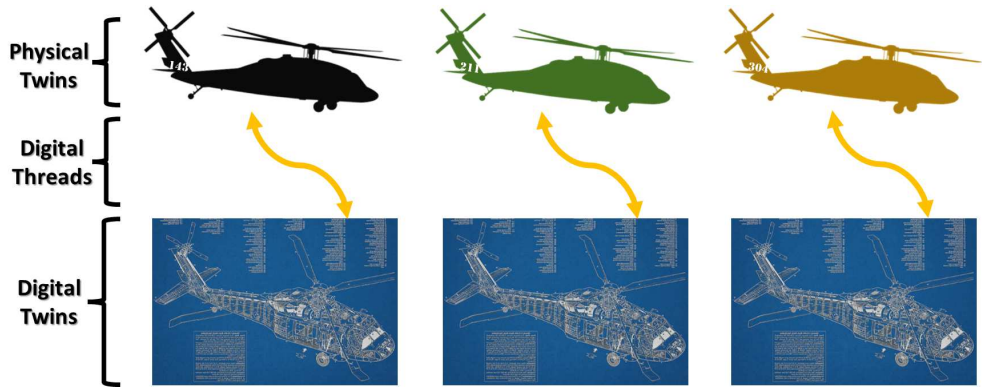}
    \caption{Illustration of a fleet of digital twins. Digital threads connect each individual physical twin with its corresponding digital twin allowing for bidirectional data and information flow.}
    
    \label{fig:DT_Fleet}
\end{figure}

The relation between the concepts of digital twin and digital thread are further illustrated in Figure \ref{fig:DT_Fleet}. A digital twin may be created for each vehicle/asset in a fleet of vehicles/assets and fleet data analyzed by the digital twins may be leveraged into information for monitoring fleet health, decreasing downtime, managing inventory, optimizing operations, and performing simulation scenarios that study performance of the fleet. 

\subsubsection{Minimally Viable Digital Twin Framework}
\begin{figure}[ht]
    \centering
    \includegraphics[width=0.8\textwidth]{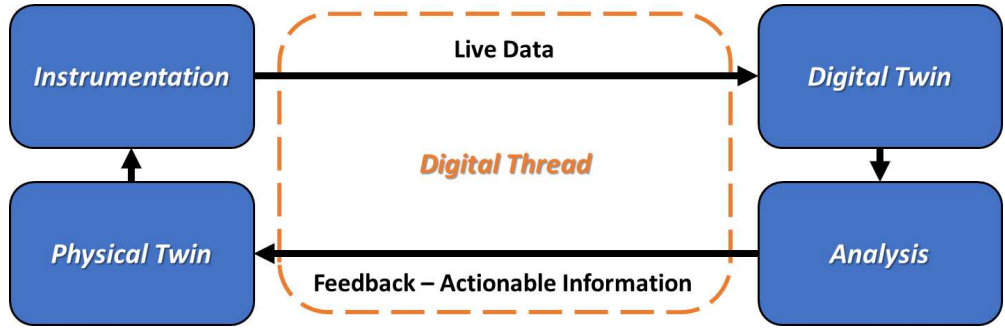}
    \caption{Minimally viable framework for a digital twin model. “Live data” is defined as a changing quantity over time with as little data capture latency as needed to allow for the information to be actionable in operations, which in turn, provides feedback on the use, control, maintenance, and/or use scheduling of the physical twin asset.}
    
    \label{fig:DT_Min_Framework}
\end{figure}

Now armed with a list of defined concepts, the task of answering, {\em what constitutes a minimally viable digital twin framework}, can be addressed. At bare minimum, a DT modeling framework (Figure \ref{fig:DT_Min_Framework}) must comprise the seven elements described below. It is worth noting that since the modeling framework may be implemented in stages, through the life-cycle of the asset, not all of the elements of the framework will be present in every stage (e.g., the physical asset may first be digitally designed) and can be considered as existing in the interim prototype stage until all elements are incorporated. 
\begin{itemize}
\item Physical Twin Asset - object, system, or process occurring in the real world; without a corresponding physical twin, the digital twin model is simply a traditional exploratory simulation or design model.
\item Digital Twin - a digital representation of the physical twin asset, such as information models describing the asset's form, function and behavior, the current state/configuration of the physical asset, as well as pointers to a digital database of historical data from instrumentation and previous states/configurations. Without the digital twin representation, the physical twin merely exists with no data and models for analysis.
\item Instrumentation – sensors, detectors, and measurement tools that collectively generate digital data and knowledge about the physical asset or the environment in which it operates; instrumentation may be independent from or embedded on the physical asset and may even include manual inspection notes converted digitally for the digital twin to interpret and process. Without some sort of instrumentation, nothing is measured and no data is generated within the modeling framework.
\item Analysis - data analytics, computational models, simulations, algortihms, and decisions (human or machine); Analysis transforms digital twin data from/of the physical asset into information that is manually, automatically, or autonomously actionable. Without analysis, the digital twin simply mirrors the current state of the physical counterpart, with no actionable information generated for the physical twin.
\item Digital Thread – a digital connection (through wired/wireless communication networks) that provides a data link between the physical asset and the digital twin. Without a digital thread, there is no digital communication between the twins, and thus the configuration comprises, at best, an analog model.
\item Live Data – data from instrumentation that is streamed through the digital thread that changes over time indicating the changing state or status of the physical asset throughout its life cycle; {\em live} being defined as the maximum amount of capture latency that allows for the feedback to be actionable. Without live data, the model is static, temporary, or non-dynamic; such a model can only provide instantaneous, temporary  solutions or prototypical off-line feedback. 
\item Actionable Information – information generated from the analysis and relayed as an actionable feedback on the physical asset. (i.e., providing feedback on the operation, control, maintenance, and/or scheduling of use of the physical twin asset). With no actionable information being utilized by the physical twin, the physical twin {\em flies blind} with no external insights into its own state or condition. 
\end{itemize}

With this framework, one can quickly distinguish what makes a digital twin model distinct from traditional design models and simulations: design models and simulations are used as testing grounds to create at least a physical asset prototype using simulated physics and computer aided design, whereas a digital twin model utilizes data directly from the field in the precise operating conditions and environment the asset is used. In addition, the authors believe the source of confusion in the multiple definitions of digital twin is alleviated by a clear distinction between digital twin and a digital twin modeling framework; the former is considered to be a digital representation of the states, conditions, environmental exposures, and configurations of the physical twin asset, while the latter (as described in the above system of seven components) provides the contextual framework in which the former is utilized. In this context, the digital twin modeling framework, which utilizes digital threads, is consistent with the idea of a {\em Cyber-Physical System} (CPS)~\cite{uhlemann_digital_2017-1}. The modeling philosophies are slightly different, though. In the CPS literature, emphasis is placed in the analysis/control tasks, and the digital representations used to assimilate instrumentation data and perform analysis (i.e., the digital twin) are typically control-oriented models of the physical asset purposefully designed to capture only the relevant behavior and properties  needed to support those specific analysis tasks. The digital twin modeling framework, on the other hand, places less emphasis on the analysis tasks and more emphasis on ensuring that the digital twin is a faithful replica of the physical asset, supporting a variety of analysis task even beyond those that are originally envisioned during the design.

A potential critique could be that this framework can be used to describe the relatively simple control logic algorithms (e.g., a vehicle cruise control or auto-pilot control) of a physical asset, or even a simple viewable dashboard of gauges that a human can view and interpret for informative feedback to control the asset. However, these simple examples miss the point that digital twins are meant to model complex interactions (multiphysics) to control complex assets (multiscale multisystems and multiprocesses). Ultimately, that is a minimal framework, and complexity implemented in digital twins can be envisaged to rise in sophistication along three main scales:
\begin{itemize}
\item {\bf Extent of autonomy and decision speed}: A basic model would be a manual supervisory one providing information to personnel who interprets the digital twin information and makes control decisions, while data or operation controls are uploaded/downloaded manually with decisions made on a coarse time scale of hours and days. An intermediate model would automate not just digital data processing but also simple stable controls with regular human interventions to mitigate unstable decision making with decisions made on the time scales of minutes to hours. An advanced autonomous model would automate the majority of decision making in real time (perhaps with only a few seconds of processing lag), instilling enough confidence to make human intervention and decision making rare.
\item {\bf Extent of component granularity}: A basic model would treat the asset as a single item, or perhaps concurrently consider a few components in isolation, and only allow for simple alarming or warning that a maintenance check by a human is needed for further investigation and troubleshooting. An intermediate model would separately model all major subsystems (e.g., avionics, engine, power equipment, hydraulics, structural frame, etc.) or known frequent problem areas, in an attempt to isolate expected faults and more easily identify anomalous behavior of major components and subsystems. An ultimate digital twin would model all components individually that make up the physical asset, as well as their potentially complex interactions in a holistic manner, living up to its "twin" title while allowing for extensive root cause analysis and adjudication of failures and faults down to individual components. 
\item {\bf Extent of incorporated physics}: A basic model would focus on a single physics domain mode of failure such as shearing of a bolt due to excessive stress. An intermediate model would incorporate at least two domains of physics to identify failures dependent on multi-domain solutions (e.g., thermal cycling leading to fatigue driven structural failure). A high fidelity model would incorporate all relevant physics domains that can cause failures (e.g., a battery explosion involving electrical, chemical, thermal, and structural considerations).
\end{itemize}

The extent of component granularity and incorporated physics included are proportional to the amount and types of instrumentation (sensor) data that is used to monitor the health and upkeep of the physical twin asset. If there are relatively few and sparsely distributed sensors and detectors in, on, or around the physical asset, then there is a higher likelihood that the digital twin models may be attempting to solve an ill-posed problem: one where unique solutions do not exist or at least where solutions are not convergent at all or that converge to local optima only. One cannot hope to achieve accurate physical simulation of all or a large percentage of the components of a complex asset without instrumentation density that captures behavior of those components at the required granularity and level of detail. Similarly, instruments should be able to capture the relevant physical quantities that the physics model attempts to represent. As an example, a digital twin thermal cycling fatigue analysis model will have no input (and no solution) if there are no embedded temperature measuring devices capturing the real environmental exposure of the components attempting to be modeled. Therefore, accurate and useful digital twin models will directly rely on sensitive, reliable, and physically relevant instrumentation. Even with entirely data-driven statistical and machine learning techniques that do not explicitly model the physics of the asset, one can expect more reliable models if training data from a set of relevant physical sensor types are included, to avoid learning spurious correlations.

\subsection{Digital Twin Literature Review of Reviews}
Many digital twin reviews exist. A recent Web of Science search for the title or abstract containing (("digital twin" OR "digital twins" OR "digital thread" OR "digital threads") AND ("review" OR "reviews" OR "survey" OR "surveys" OR "landscape" OR "landscapes")) returned 215 results. A similarly filtered Google Scholar\texttrademark{} search limited to review articles returned 305 results. A comprehensive review of digital twin review articles is in order, but it is beyond the scope of this manuscript. We only provide a brief overview with highlights below.

Lu et al.~\cite{lu2020digital} is the most cited review of the digital twin concept. The authors provide a systematic examination of the definition of digital twin and related terms, the enabling technologies, communication standards and network architecture of the digital thread, and  current research issues and challenges. Nevertheless, the review is singularly focused on industrial manufacturing applications. Also, instead of demonstrating specific examples of DT, general advantages, such as "decision-making optimization", "data sharing", and "mass personalization" are outlined without quantitative metrics in support. Kritzinger et al.~\cite{kritzinger2018digital} is another frequently cited review of digital twin use in industrial manufacturing. The authors provide a classification schema of DT relevant articles based on the level of automation and model types defining terms such as "digital model" (manual data flow between physical and digital objects in both directions), "digital shadow" (automated data flow from  physical to digital objects, but manual data flow from digital to manual objects), and "digital twin" (automated data flow between physical and digital objects in both directions)~\cite{kritzinger2018digital}. Only about one quarter of cited articles involved case studies. 

Errandonea et al.~\cite{errandonea2020digital} provide a comprehensive review of digital twins used in the context of predictive maintenance. The paper identified 68 articles and conference papers that used DT for maintenance applications. Khan et al.~\cite{khan2020requirements} provide a look at the path towards {\em autonomous maintenance}, which explores requirements from sensors, machine learning, security, and material science that are needed in order to achieve highly automated and low human intervention digital twin models for equipment maintenance.

Khan et al.~\cite{khan2018review} is another comprehensive, as well as visionary, review of deep learning applied in the context of system health management (SHM). SHM includes PMx, yet is a more general term, encompassing diagnostics and anomaly detection in addition to the prognostics often performed in PMx. The authors identified 38 articles that use deep learning methods to analyze and interpret data from equipment to perform anomaly detection, fault and failure diagnosis and classification, remaining useful life estimation, and component degradation.

Rasheed et al.~\cite{rasheed2020digital} outline several hybrid methods for digital twins incorporating ML and Physics-Based Modeling (PBM); however, there are no current reviews on the intersection of digital twin predictive maintenance models incorporating both machine learning methods and physics based modeling. 

\section{Digital Twin Applications and Industries}
We now review possible digital twin applications and the examples from different industries, in order to better illustrate how the concepts described so far map into real-world usage scenarios.

\subsection{DT Applications}
We begin with a review of prototypical applications that are relevant to multiple different industries.

\paragraph{Risk assessment and decreased time to production:}
For manufacturing and production, a plant layout is a time intensive design and high cost endeavor, which inherently involves risk from inefficient allocations of time and resources. The digital twin creates an opportunity to decrease the time to production by effectively creating a high fidelity representation of the production line in a virtual space, while also exploring what-if design scenarios that can result in optimization of layout to maximize production output and/or minimizing costs~\cite{negri2017review}. This virtual simulation of the manufacturing line may also be used determine what variables of the system are important to monitor in the operation phase and further, whilst in the operation phase, the virtual digital twin, used initially for design, may then be updated with sensor data for the algorithms to analyze and produce prognostics for operations and maintenance. 

\paragraph{Predictive maintenance:}
Predictive maintenance has been stated as both the original purpose for the concept of digital twin \cite{shafto2012modeling} as well as being the most popular application of the digital twin model \cite{errandonea2020digital}. The predictive maintenance application will be discussed in further detail in Section 3.

\paragraph{Real-time remote monitoring and fleet management:}
After the predictive maintenance models arrive at their updated predictions, fleet management may then be optimized and executed. Fleet management involves maintenance planning, use scheduling, and logistics as well as performance evaluation metrics. Verdouw et al.~\cite{verdouw2017digital} summarizes several digital twin applications in agriculture, including a fleet management application that tracks individual equipment location and energy use, accurate row tracking with various towed agricultural equipment, and evaluation of crop yield for individual machines. Major et al. provides a demonstrated example of real-time remote monitoring and control of a ship's crane, as well as the ship itself, with the long-term goal of improved fleet logistics and safety, and supervision from onshore monitoring centers~\cite{major2021real}.

\paragraph{Increased team collaboration and efficiency:}
Efficient collaboration between members of a project team is vital if a project is to stay within time and budget constraints. Data and information from assets, as it is updated, altered, and generated must be shared with project managers, multidisciplinary engineering groups, builders and/or manufacturers, and customers/consumers. The need for an integrated platform is substantial, especially when team collaborators are of various technical and skilled backgrounds. The digital twin framework provides such a platform that potentially allows near real time monitoring and information interrogation from a consolidated reliable source. Perhaps this is best illustrated in the field of construction; Lee et al.~\cite{lee2021integrated} disclose a blockchain framework for providing traceability of updates to the database that warehouses both the {\em as-planned} construction and the {\em as-built} project generated from GPS as-measured geometry, and material property testing (e.g., soil and building material). The blockchain traces and authenticates user updates that become a single real time source of truth for all users to interrogate. One of the biggest sources of delays in construction projects occurs from lack of adequately planning for supply chain logistics: getting the right building materials to the right place at the right time. Having a reliable and authenticated source of project status could allow optimization of supply orders and deliveries.

\subsection{DT Industries}
We now review specific applications within different industries.

\paragraph{Manufacturing:}
Digital twins have been extensively applied in industrial manufacturing, mainly in the form of a predictive maintenance model of large complex manufacturing machinery and in the form of a large scale simulation of production run machinery, the latter having the goal of decreasing time to production and assessing risk as is mentioned above. Uhlemann et al.~\cite{uhlemann_digital_2017-1} give an example of effective and efficient production layout planning for small to medium-sized enterprises where different generated layouts are compared using a digital twin simulated environment. PMx applications for manufacturing are also a target area for DT model application, focusing on the impact of upkeep on output production as well as correlated maintenance on upstream and downstream equipment~\cite{susto2014machine}.  Rosen et al.~\cite{rosen_about_2015} explore production planning and control by developing a DT simulation framework to optimize the effects of production parameters on output production and manufacturing equipment maintenance.

\paragraph{Aerospace:}
The digital twin framework is heavily applied in aerospace and aviation in predictive maintenance and fleet management applications~\cite{glaessgen_digital_2012,karve_digital_2020,musso2020interacting,sisson2022digital,wang2020life,xiong2021digital,zaccaria2018fleet,zhou2021real,kapteyn2020toward,guivarch_creation_2019}. The large amount of sensor recorded data and up-to-date maintenance records create an environment in which predictive maintenance models can thrive, but also necessitates large scale data management to provide authoritative and easy-to-interrogate databases. Simulation studies evaluating performance, faults, or failure of assets utilizing digital twin models may be performed for various operating conditions of the assets and also under various ambient environmental conditions. Logistical plans may be developed and explored based on simulated site plans or various fleet sizes~\cite{west2018demonstrated}. Machine learning algorithms are employed to flexibly optimize logistical problems associated with predictive maintenance while also predicting the faults and failures of assets based on trends and patterns in the recorded sensor data. These faults may be identified from anomaly detectors that make use of recorded {\em healthy} or {\em nominal} operational data that contrast with anomalous data, which can be flagged in real time. 

\paragraph{Architecture, Civil Engineering, Structures and Building Management:}
Building information models (BIM) are closely related to digital twins and are very well known in the architecture, engineering and construction management world. BIMs are semantically rich information models of buildings that can be used to easily visualize the 3D representation of the building along with key properties of the different components and systems. They have also been used to assimilate changes made to the physical building, and in turn, act as a guide during building design, construction and maintenance operations~\cite{coupry2021bim}. Structural health modeling is also a salient example of digital twin use for maintenance, repair and operations (MRO) of structures ~\cite{bigoni2022predictive,droz2021multi,taddei2018simulation, rosafalco2020fully}. Here, digital twins incorporating both analytics of data-driven machine learning of states attributed to sensor data and physics based models, which provide simulations for training models, work together to provide warnings and predictions of potentially adverse states of the structure.

\paragraph{Healthcare, Medicine, and the Human Digital Twin:}
In clinical settings, one is often surrounded by various complex and expensive equipment which is also the subject of maintenance, repair, and fleet management. Healthcare costs continue to rise, yet the high costs of preventive maintenance still dominate healthcare environments, where new technologies can be adopted slowly~\cite{shamayleh2020iot}. There is potential for large cost savings by adapting digital twin based predictive maintenance models. The digital twin concept has also been applied directly to the human patient as well with high fidelity cardiac models serving to help diagnose heart conditions that may lead to patient specific and personalized treatments~\cite{gillette2021framework}. \cite{corral2020digital} Yet another healthcare application is that of a digital twin model of humans in the context of wearable health monitors and trackers that collect data and provide personalized advice, diagnoses, and treatments based on data-driven prediction models~\cite{liu2019novel}. Along similar methodology, Barbiero et al.~\cite{barbiero2021graph} use a patient digital twin approach utilizing a graph neural network to forecast patient conditions (e.g., high blood pressure) based on clinical data from multiple levels of anatomy and physiology, such as cells, tissues, organs, and organ systems.

\section{Interrelation between Predictive Maintenance and Digital Twin Framework} 
\subsection{Distinguishing Types of Maintenance}
\begin{figure}[ht]
    \centering
    \includegraphics[width=0.4\textwidth]{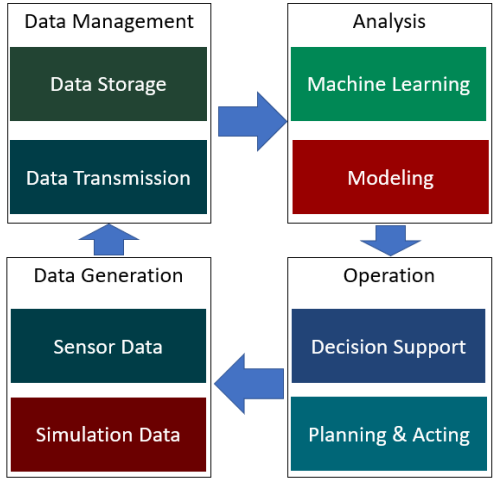}
    \caption{The layers of the predictive maintenance stack fit together in a control loop where data is generated, transmitted, analyzed, and acted on. This figure is reproduced with permission from Edman et al.~\cite{edman2021predictive}}
    
    \label{fig:pmx-control-loop}
\end{figure}

Predictive maintenance and the digital twin modeling framework share many common elements and goals. Predictive maintenance  utilizes sensor data or maintenance history data from a physical asset to predict the probability of failure over a future interval. The prediction may be used to plan and execute future maintenance or operation of the physical asset. This cycle can be visualized as a control loop, (Figure \ref{fig:pmx-control-loop}) wherein data is generated from the instruments associated with the physical twin asset, passed through the digital thread via the data management block, analyzed by the digital twin models/algorithms, turned into actionable information to operate the physical twin asset, which then leads to new data generation, beginning the cycle anew. Thus, predictive maintenance models, when deployed, meet all the criteria of the definition of a minimally viable digital twin framework (Figure \ref{fig:DT_Min_Framework}). Therefore, it is often mentioned that predictive maintenance models are one of the most popular applications of digital twin~\cite{errandonea2020digital}. However, in the broader field of maintenance, repair, and overhaul (MRO), several distinct maintenance types exist depending on the desired level of safety, cost savings and available instrument data, and not all maintenance operations are fit to utilize a digital twin approach. 

\paragraph{Reactive Maintenance:}
Reactive maintenance may be summarized by the adage, {\em if it isn't broke, don't fix it}. Equipment is allowed to run to failure. Failed parts are replaced or repaired. Repairs may be temporary in order to quickly regain operational status, while putting off until later more time consuming permanent repairs. Such an approach may save money and personnel time in the short term, but runs the risk of unpredictable interruption and more costly catastrophic failure in the future~\cite{swanson2001linking}. No data from the asset, even if any is recorded, is analyzed, therefore this approach does not utilize the digital twin framework.

\paragraph{Preventative Maintenance:}
Some assets cannot risk catastrophic failure from the perspective of both a total loss/destruction of the asset and loss of human life (i.e., such as catastrophic failure of an aircraft). Instead, the goal of preventive maintenance, also known as scheduled maintenance, is the exact opposite of the reactive approach, in that safety is of utmost importance. In preventive maintenance, there is little tolerance for risk of failure in any critical components. Maintenance plans are set by manufacturers with scheduled inspections and overhauls to occur well before estimated time of failure utilizing large safety margins~\cite{shafiee2015maintenance}. However, such precautions come with the penalty of prematurely replacing still usable components. Again, since no asset data is analyzed this approach does not utilize the digital twin framework.

\paragraph{Condition-Based Maintenance:}
Condition-based maintenance (CBM) falls between the {\em never replace before failure} philosophy of reactive maintenance and the {\em always replace before there is a remote chance of failure} philosophy of preventive maintenance. CBM can be synonymous with the term {\em diagnosis based maintenance} and is sometimes used interchangeably with PMx or with the term {\em condition monitoring}, although there is not a consensus on the equivalency of these terms, which is a source of confusion~\cite{si2011remaining}. However, CBM may have a subtle distinction from PMx, in that CBM requires a keen human observer and tends to use human senses (e.g., odd noises, smells, vibrations, visible erratic behaviors, etc.) to sense that the condition of the asset is in an anomalous state~\cite{nikolaev2019hybrid}. Other CBM methods could be simple heuristics applied to sensor data. An example heuristic would be accelerometer measurements to not exceed the threshold of $k$ standard deviations of signal from some past defined interval, otherwise an impending failure is expected. CBM is used to assess the state of the asset in the present moment based on a recent past interval of time and thus is typically not considered a predictive method. CBM rules are often static and sensitive to noise and artifacts that violate assumptions but do not increase risk of failure, which may lead to excessive maintenance~\cite{errandonea2020digital}. CBM methods can fall under the category of the digital twin framework if they make use of digital data, but often fail to meet the criteria, due to a lack of digital thread, if only analog signals are analyzed and actionable feedback is manually implemented.

\paragraph{Predictive Maintenance:}
Predictive maintenance is a proactive technique that uses real-time asset data (collected through sensors), historical performance data, and advanced analytics to forecast when asset failure will occur. Using data collected by sensors during normal operation, predictive maintenance software uses advanced algorithms to compare real-time data against known measurements, and accurately predicts asset failure. Advanced PMx techniques incorporate machine learning, which is summarized in several extensive reviews~\cite{carvalho2019systematic,miller_system-level_2020}. The result of PMx is that maintenance work can be scheduled and performed before an asset is expected to fail with minimal downtime or impact on operations~\cite{nikolaev2019hybrid}. PMx is synonymous with the the terms prognostic health management (PHM) and systems health management (SHM)~\cite{khan2018review}. As outlined above, predictive maintenance algorithms and models are an application of the digital twin modeling framework.

\paragraph{Prescriptive Maintenance:}
Prescriptive maintenance can be thought of as the implementation step after predictive maintenance~\cite{errandonea2020digital}. After an asset or fleet of assets are predicted (often with a probability or confidence window) to fail within a time interval into the future, the next task is to optimize the maintenance schedule that minimizes costs, minimizes equipment downtime, and maximizes logistical efficiencies (getting the right replacement component in the right place at the right time). Prescriptive maintenance is essentially the optimization and execution of the maintenance plan after predictive maintenance has been performed. Note that the concept of {\em prescriptive maintenance} is often included under the term {\em predictive maintenance}.  

\paragraph{Diagnostics, Anomaly Detection, and Prognostics: What Went Wrong?, What Is Going Wrong?, What Will Go Wrong?}
Maintenance, repair, and operations is often divided into the distinct tasks of diagnostics, anomaly detection, and prognostics. ~\cite{khan2018review} 

Diagnostics is the task primarily concerned with classifying anomalous behavior to known fault conditions (i.e., to answer the question of: what went wrong?). Faults, or the erroneous operation of equipment, are identified using a diagnostic framework. Anomaly detection is chiefly concerned with detection of unintended or unexpected functions of the monitored equipment. Anomalies may or may not cause or lead to faults or failures, but simply are significant deviations from nominal operation as recorded in the past. Ideally, all potential anomalies and diagnoses would be accounted for during the design and testing phase of the equipment development, yet in practice, complex or aging assets can fail in unanticipated modes and scenarios and therefore, in-the-field analysis of detected anomalies and their linking to discovered faults is at the heart of digital-twin data-driven techniques for diagnostics. Root cause analysis (RCA) is the systematic approach for diagnosing the fault (cause) that leads to a failure. Traditional RCA approaches have included manual analysis by subject matter experts (SMEs) using diagnostic fault trees, fishbone diagrams, and the 5 whys procedure~\cite{sivaraman2021multi}. However, these manual analysis methods become unwieldy with large systems that have many levels of component interaction. As equipment becomes more complex and sophisticated, the number or combinations and permutations of potential causal factors for certain fault events rapidly increases. Therefore, statistical tests and analytic methods, such as regression and the Pearson correlation coefficient, have been applied to capture relationships between recorded sensor variables~\cite{madhavan2021evidence,zhao2017advanced}. Nevertheless, even these methods have difficulty in dealing with non-linear patterns, as well as multi-variable dependence effects and multiple timing and lag effects~\cite{bonissone2009systematic}. 

Finally, the task of prognostics deals with predicting a future state or condition (e.g., failure) of the equipment or component thereof. The prediction comes with uncertainties (or are probabilistic) that typically expand with the span of the forecast horizon. Remaining useful life (RUL) estimations of equipment components, effectively a time-to-failure prediction, are the most common type of prognostic indicator and traditionally, have been calculated using historical operational component data (e.g., similar components utilized across a fleet of assets) analyzed with statistical techniques such as regression, stochastic filtering (e.g., Kalman filter), covariate hazard (e.g., Cox proportional hazard), and hidden Markov models~\cite{si2011remaining}.

\section{Incorporating Machine Learning with the Digital Twin Framework for Predictive Maintenance}
Machine learning (ML) is a broad field of various analytical methods that learn by leveraging historical data to make decisions about data encountered in the future. Accordingly, since predictive maintenance, an application of digital twin models, utilizes instrumentation data to make diagnostic or prognostic decisions, ML has been often applied to PMx analyses~\cite{errandonea2020digital,khan2018review,nikolaev2019hybrid}. ML algorithms can be coarsely divided into three different types of learning: supervised, unsupervised, and semi-supervised. These different types are separated based on the type of data one can employ; supervised learning works with data that has labels (i.e., the true predictive outcome is linked to each observation); unsupervised learning has no access to ground-truth labels an utilizes techniques to group, cluster, or extract patterns that can be distinct indicators of an underlying phenomenon; while semi-supervised learning is a hybrid approach in which only a small portion of the data has labels and it may be possible to infer the missing ground-truth annotations. 
Likewise, in PMx applications, the approach chosen will depend on the instrumentation data one can use and whether the data is labelled, that is, faults and failures, as well as nominal operation patterns, have been attributed to the recorded data. Ideally, data is collected from multiple sensors over a fleet of assets, over the entire life of each component, from initial normal operation, through a time of degradation, and then finally until failure. Despite increasing use of connected sensors embedded on complex assets, access to these ideal databases can be rare due to the storage requirements needed to warehouse prolonged acquisition data cycles. Also, many complex physical assets are rarely run until failure, due to the risk of a catastrophic consequences, including total loss of the asset. 

\paragraph{Predictive Maintenance Workflow and Training from Nominal {\em Healthy} Data}
Booyse et al.~\cite{booyse2020deep} argue that the first phase in health monitoring is to be able to detect anomalous or faulty behavior from nominal healthy behavior. Typically this involves a workflow of data acquisition (DAQ) from the instrumentation, preprocessing of data to filter noise and remove artifact, identifying a set of distinct condition indicators that differentiate normal and anomalous data or previously known fault modes, training and testing an ML model to enable predictions on test data, and deploying the model to make predictions when encountering future data. Often, the acquired data is time series data (i.e., a recorded numeric variable, such as temperature, pressure, acceleration, concentration, strain, etc., as a function of time) and condition indicators, or features, can be based in the time-domain (e.g., mean, standard deviation, root-mean square, skewness, kurtosis, or morphological parameters of the shape of the time series signal, etc.), frequency-domain (e.g., power bandwidth, mean frequency, peak frequencies and amplitudes, etc.) and time-frequency domain (e.g., spectral entropy, spectral kurtosis, etc.). Condition indicators could also be derived from static or dynamic physical models. Determination of anomalous behavior may involve one or more distinctive features which may be extracted by unsupervised learning methods, such as clustering, or if many features are involved, dimension reduction maybe performed through, e.g.,  principal component analysis or other null-space basis model type. Yang et al.~\cite{yang2022causal} describe a causal correlation algorithm that is applied to Bayesian networks and potential for diagnosing causality of behaviors (e.g., faults) in large complicated networks that have non-linearities and are multivariate. As faults are identified and accumulated during the operational history of the asset, these events may then be compared to the identified groups or clusters, thus performing a diagnostic function of linking anomalous behavior with identified faults.

\paragraph{Similarity Models: When Data Encompasses Run to Failure Complete Histories}
If the acquired data is a complete history of a group (fleet) of similar equipment, spanning from its initial operation until failure, it lends itself to prognostication of failures using similarity models~\cite{Mikus2007},~\cite{Dubrawski2011}. 
If one can map the current state of an asset on the specific time marks of usage trajectories observed from other assets whose actual outcomes are known, one could use the distribution of these outcomes as a reference for e.g., remaining useful life, time to specific type of failure, or other statistics of interest. Many of similarity based modeling techniques draw inspiration from medicine, where the individual patient's heath status assessment, diagnosis as well as prognosis are routinely mapped on the background of many similar cases observed before, sometimes including detection of distinct phenotypes of trajectories of evolution towards failure~\cite{Chen2015}, or from public health, where the task of detecting outbreaks of infectious diseases appears similar to the task of detecting onset of new types of failures spreading across the fleet of an equipment~\cite{Dubrawski2007}. Applications of this concept vary in how the similarity is defined and how the predictions are formed, and include statistical machine learning as well as neural network based methods ~\cite{adhikari2018machine,saha2019different,bektas2019neural}.

\paragraph{Survival Models: When Data Encompasses Only the End Time of Failure From Similar Equipment}
Sometimes only the time point of failure is known, and survival models such as Kaplan-Meier, Cox proportional hazard (CPH)~\cite{hrnjica2021survival}, or more advanced approaches using deep learning~\cite{chen2020predictive} can be utilized. Survival curves (probability of survival over time) for time-to-event (i.e., events are faults or failures in the PMx context) are generated from the failure history data. Distinct survival curves may be generated for groups of equipment that share similar covariates (similar condition indicators or properties, e.g., manufacture, operating conditions). Based on the survival curves, RUL may be estimated~\cite{hrnjica2021survival}.

 \paragraph{Degradation models: When Data Encompasses Run to Known Threshold that Exceeds Safety Criteria}
 Frequently, equipment is not run until failure, but instead until just before exceeding a safety threshold. A class of methods that reflect this concept is known as degradation models. They can be implemented as a linear stochastic process or an exponential stochastic process if the equipment experiences cumulative degradation effects~\cite{thiel2021cumulative}. Degradation models typically work with a single condition indicator, although data fusion techniques can be used to combine multiple indicators of degradation. However, what role maintenance and repair may have in not only censoring failure events, but also resetting or offsetting latent degradation states is largely an unexplored area of research~\cite{miller_system-level_2020}.   

\section{Incorporating Physics-Based Modeling and Simulations with Digital Twin Frameworks for Predictive Maintenance}

So far, when discussing PMx tasks and their specific applications, the digital twins at the core of the digital twin framework have been largely statistical models of the behavior of the physical asset, tailored specifically to address the PMx task in question.  However, any careful reader may have observed already that there is ample opportunity to extend the reach of the process by leveraging more complex (e.g., multi-physics, multi-scale) models of the assets being twinned. Physics-based models (PBMs) are extensively utilized in the design-phase and are excellent candidates for such extensions. However, a drop-in replacement may be difficult as digital twins and PBMs differ in several fundamental ways:
\begin{enumerate}

\item PBMs are often used as a design tool in the early phases of the product or system development and are rarely returned to after the design, testing, and manufacturing phases are completed, perhaps only after product failure (failure analysis). Digital twins are updated and utilized through all phases from concept, design, and testing to implementation, customer support, and end of product life phases. Ensuring that PBMs can be easily updated throughout the life-cycle requires some adjustments.

\item Traditional PBMs are often built only for specifically assumed operating conditions for the product or the conditions are learned from limited bench-top lab testing or environmental data. The digital twin modeling framework relies on generated data from real world applications of products; simulations and analyses performed with the use of digital twins are enabled with data from the fully assembled, working product or system, operating in their real-world environment. Thus, continuously calibrating/update PBMs to enable this integration can pose a challenge.

\item Traditional PBMs are typically highly detailed representations of small parts or subsystems of an overall assemblage of parts or a system-of-systems. Digital twin analyses take advantage of data being collected from the working assembled system to predict faults and failures that occur due to the complex interactions of systems of parts and their environment. PBMs typically require orders of magnitude more computations to produce their predictions than the data-driven models typically used in PMx. 

\end{enumerate}

It would be unfair to characterize the relationship of traditional PBMs and digital twin models as adversarial. Due to the benefits they provide, many of the design phase PBMs already form the initial record of the digital twin, or of the individual parts that comprise a digital twin of a system or asset. We now expand on the specific limitations of using PBMs in the digital twin modeling framework.

\paragraph{Limitations of Multi-Scale, Multi-Physics Models for the Digital Twin Framework}
A common refrain from digital twin skeptics may be that a realistic digital twin can never fully simulate a physical system or asset; however all models, including physics-based models (e.g., stress/strain analysis, computational fluid dynamics, electromagnetic scattering, and other complex processes) are based on assumptions and simplifications in an attempt to explain complex physical interactions. Yet, despite the simplifications, computational models have driven much of engineering progress over the past half century. PBMs often achieve accurate solutions by discretizing, splitting the volumes or areas into small regular elements and nodes that populate a model space governed by differential equations. Typically, these are second order (or higher), non-linear, partial differential equations, which have few, if any, general solutions or techniques. If the model space is of irregular geometry as well, a discretized model based on finite or boundary elements may be the only option for a reasonably accurate solution. Unfortunately, the accuracy of such models come at the cost of extensive computation time. Cerrone et al.~\cite{cerrone2014effects} estimated a single simulation run of crack propagation in a structural plate with several notches and holes had: 
\begin{quote}
approximately 5.5 million degrees of freedom. Simulations were conducted on a 3.40 GHz, 4th generation Intel Core i7 processor. Abaqus/Explicit’s shared memory parallelization on four threads with a targeted time increment of $1\cdot10^{-6}$ seconds resulted in approximately a 4-day wall-clock run time~\cite{cerrone2014effects}.
\end{quote}

Of course, such a runtime is far from the promise of digital twins providing near real-time decision making from the constant stream of sensor data. To exacerbate, modern automobiles have tens of thousands of parts and large commercial aircraft may have millions of parts~\cite{airbus_a380_facts}. In addition, engineers frequently desire more than one type of {\em what if} simulation and would prefer to run a great number of varied simulations to explore variables and effects. These complicating factors would seem to relegate the concept of digital twins based entirely on physics models out of reach practically and economically~\cite{west2017digital,tuegel2012airframe}, or at least push its utility decades into the future. However, there are several novel forms of analysis that seek to hybridize physics based models with  purely data-driven techniques, such as machine learning, which may result in more manageable computational costs of PBMs and automated learning of their parameters.

\paragraph{Combining Data-Driven Machine Learning Methods and Physics-Based Models to Address Each Other's Shortcomings}
While a digital twin definition often includes the concept of physics based models, there are well known limitations of such models as outlined above. The term {\em model} may be generalized to include trends and patterns directly learned from data. 
For example, there is the related terminology of implicit digital twins (IDT) from Xiong et al.~\cite{xiong2021digital}:

\begin{quote}
However, the traditional DT method requires a definite physical model. The structure of the aero-engine system is complex, and the use of a physical-based model to implement a DT requires the establishment of its own model for each component unit, which complicates predictive maintenance and increases costs, let alone achieve accurate maintenance. To circumvent this limitation, this paper uses data-driven and deep learning technology to develop DT from sensor data and historical operation data of equipment and realizes reliable simulation data mapping through intelligent sensing and data mining (called implicit digital twin; IDT). By properly mapping the simulation data of aero-engine cluster to a certain parameter, combined with the deep learning method, various scenarios’ remaining useful life (RUL) can be predicted by adjusting the parameters.~\cite{xiong2021digital}
\end{quote}

\noindent In other words, a physical model may not be needed; an IDT model can simply be created through data-driven analysis (i.e., machine learning). Nevertheless, ML approaches have a few relevant limitations as well:
\begin{enumerate}
\item Scientific and engineering problems are often underconstrained (i.e., large number of variables, small number of samples) making learning reliable ML models from the corresponding data difficult.
\item Catastrophic failures are naturally infrequent and so they may be seldom, if ever, encountered in the recorded data. This issue can sometimes be alleviated using Bayesian forms of ML that allow incorporation of prior probability distributions to account for events that are not represented in data.
\item It can be easy for crossvalidation methods to misevaluate spurious relationships learned by data-driven frameworks as they can look deceptively well on training as well as tests sets. 
\item Some ML methods, such as deep neural networks, are ``black boxes'' that provide few interpretable insights into the resulting models, and as such they may fail to convince their users that the obtained solutions are sufficiently systematic to be applicable to future, similar problems.
\item Data used for training ML models is most often just a limited projection of the reality that they are expected to capture. It is then easy for even advanced ML models to fail to follow common sense that comes natural to human domain experts, if important nuances of the underlying knowledge is not reflected in the training data.
\item Large amounts of labeled training data that is often required to produce reliable data-driven models can be expensive to acquire and/or time consuming to create.
\end{enumerate}

Some recently developed frameworks aim to combine the interpretability of PBMs with the data-driven analytical power of digital twins and machine learning. 
Combining PBMs and ML seeks to overcome the problems of the long runtime of highly detailed and complex physical models on one hand, and the lack of interpretability and required large volumes of labeled training data on the other~\cite{kapteyn2020toward}. Hybrid ML and physics based solutions are likely to excel in solving analytic problems, such as planning maintenance of complex assets (e.g., vehicles, aircraft, buildings, manufacturing plants) where some data (e.g., small samples of rare occurrences, noisy and spatially or temporally sparse sensor recordings, etc.) and some knowledge of physics (e.g., missing boundary conditions, occurrence of physical interactions, etc.) exists, but neither alone are likely to contain enough information to solve complex diagnostics or prognostics problems with sufficiently useful accuracy or precision for practical application. 

\paragraph{Hybrid Digital Twin Frameworks Combine Machine-Learning and Physics-Based Modeling}
There are several approaches to combining PBMs and ML, which may go by several different names, such as {\em physics informed machine learning} (PIML), {\em theory guided data science} (TGDS), {\em scientific machine learning} (SciML), which include physical constraints and known parameter relationships (e.g., an ODE, PDE, physical and material properties and relationships). There are three main strategies identified so far: physics informed neural networks, reduced order modeling, and simulated data generation for supplementing small data sets.

The work of adjusting PBMs to work well with ML (and in particular for the PMx tasks) has been so far limited to very simple components and systems. The R\&D community is yet to seriously consider scaling hybrid approaches to handle complex systems with multiple components requiring true multi-physics and multi-scale models. Work is needed on both figuring out how to automatically integrate multiple PBM models together, as well as how to automatically select the right hybridization strategy to make the resulting solutions run sufficiently fast without compromising fidelity of the resulting models.

\paragraph{Physics Informed Machine Learning (PIML) and Physics Informed Neural Networks (PINNs)} In 2019, Raissi et al.~\cite{raissi2019physics} proposed a deep learning framework that combines mathematical models and data by taking advantage of prior techniques for using neural networks as differentiation engines and differential equation solvers (e.g., Neural ODE~\cite{chen2018neural}). The main idea is that the physical relations and equations modeling them are leveraged to formulate a loss function for the ML algorithm to minimize violation of the principles of physics. 
Specifically, a loss function can combine the usual data-driven component based on observed residuals, with physics-driven terms reflective of errors in the solutions of the governing ODE or PDE, and terms reflective of violations of any boundary or initial conditions. 
Raissi et al.~\cite{raissi2019physics} provide several examples of solved dynamic as well as boundary value problems, including Schrodinger's, Navier Stokes, and Berger's equations. A similar approach by Jia et al.~\cite{jia2019physics} under the name of {\em physics guided neural networks} (PGNN) which was used to determine temperature distribution along the depth of a lake using both physical relations and sensor data.

\paragraph{Reduced Order Modeling}
Another method is to reduce the order, size, number of degrees of freedom (DoF), or dimensionality of the PBMs. This approach is often called {\em reduced order modeling} or {\em projection based modeling} or {\em lift and learn}~\cite{swischuk2019projection,kapteyn2020toward}. 
Here, training data is generated by the PBMs, but only a few snapshots, e.g., a few individually solved time instances are used to reduce the computational power needed instead of solving over a complete time domain. 
Then, a lower-dimensional basis is computed and the higher-order PDE model is projected onto the lower-dimensional space. Hartmann et al.~\cite{hartmann202012,hartmann2018model} give an excellent review on reduced order modeling and its role as a digital twin enabling technology, drastically reducing model complexity and computation time, all while maintaining high fidelity of solutions.
The reduced order model solutions can be arrived at rapidly for various boundary and/or initial conditions and are frequently used as a simulation database to which ML algorithms may be trained on to classify damage states or learn a regression to a continuous degradation model; the trained models can then be fed distributed sensor data from real-world assets in the field to perform equipment health monitoring tasks~\cite{bigoni2022predictive,kapteyn2020toward,hartmann2018model,droz2021multi,leser2020digital,taddei2018simulation,rosafalco2020fully}. 


\section{Challenges of Digital Twin Implementation}
\paragraph{Sensor Robustness, Missing data, Poor Quality data, and Offline Sensors}
DT frameworks require live data, which is often generated by a dense array of sensors. Inevitably, one or more sensors will disconnect, contain periodic noise or artifact, or ironically, require maintenance. First, the DT models need to be able to detect and handle sensor signal dropout. If not accounted for in the model algorithms, faults may go unnoticed or misdiagnosed. One such way of dealing with data interruptions is to use {\em circuit breakers} attached to the sensors that trip when the signals go out of range. \cite{preuveneers2018robust} Another approach is to ensure adequate signal processing to remove and filter out unwanted noise and artifacts.

\paragraph{Workplace Adoption of the DT Framework}
Successful DT frameworks require a team effort of caring about data quality. Errors introduced through improper data entry or inadvertent part swaps will propagate throughout the DT framework. Improving user interfaces on data entry menus, as well as seeking out and requesting feedback from team members will demonstrate concerns for the daily user. Finally, sharing the goals, the rewards, and the output of the models with team members also helps reinforce positive feedback in the workplace.

\paragraph{Security Protocols}
It is estimated that the majority of network-connected digital twins will utilize at least five different kinds of integration endpoints, while each of the endpoints represents a potential area of security vulnerability. \cite{mullet2021review} Therefore, it is highly recommended that best practices be implemented wherever possible, including, but not limited to: end to end data encryption, restricted use of portable media such as portable hard drives and other portable media on the network, regular data backups (to offline locations if possible), automated system software patch and upgrade installs, password protected programmable logic contollers, managed user authentication and controlled access to digital twin assets~\cite{mullet2021review}.

\section{Summary}
This manuscript attempts to provide clarity on defining {\em digital twin}, through exploring the history of the term, its initial context in the fields of product life cycle management, asset maintenance, and equipment fleet management, operations, and planning. A definition for a minimally viable digital twin framework is also provided based on seven essential elements. A brief tour through DT applications and industries where DT methods are employed is also provided. Thereafter, the paper highlights the application of a digital twin framework in the field of predictive maintenance, and its extensions utilizing machine learning and physics based modeling. The authors submit that employing the combination of machine learning and physics based modeling to form hybrid digital twin frameworks, may synergistically alleviate the shortcomings of each method when used in isolation. Finally, the paper summarizes the key challenges of implementing digital twin models in practice. A few evident limitations notwithstanding, digital twin technology experiences rapid growth and as it matures, we expect its great promise to materialize and substantially enhance tools and solutions for intelligent upkeep of complex equipment. 


\paragraph{Acknowledgement:} 
This work was partially supported by the U.S.\ Army Contracting Command under Contracts W911NF20D0002 and W911NF22F0014 delivery order No. 4 and by a Space Technology Research Institutes grant from NASA’s Space Technology Research Grants Program.

\bibliographystyle{plain}
\bibliography{refs.bib}


\end{document}